# Leveraging Foreign Language Labeled Data for Aspect-Based Opinion Mining

*Nguyen Thi Thanh Thuy, Ngo Xuan Bach, Tu Minh Phuong**

*Department of Computer Science and Machine Learning & Applications Lab, Posts and Telecommunications Institute of Technology, Hanoi, Vietnam*
*Emails:   thuyntt@ptit.edu.vn     bachnx@ptit.edu.vn     phuongtm@ptit.edu.vn*
*\* Corresponding author*

**Abstract:** Aspect-based opinion mining is the task of identifying sentiment at the aspect level in opinionated text, which consists of two subtasks: aspect category extraction and sentiment polarity classification. While aspect category extraction aims to detect and categorize opinion targets such as product features, sentiment polarity classification assigns a sentiment label, i.e. positive, negative, or neutral, to each identified aspect. Supervised learning methods have been shown to deliver better accuracy for this task but they require labeled data, which is costly to obtain, especially for resource-poor languages like Vietnamese. To address this problem, we present a supervised aspect-based opinion mining method that utilizes labeled data from a foreign language (English in this case), which is translated to Vietnamese by an automated translation tool (Google Translate). Because aspects and opinions in different languages may be expressed by different words, we propose using word embeddings, in addition to other features, to reduce the vocabulary difference between the original and translated texts, thus improving the effectiveness of aspect category extraction and sentiment polarity classification processes. We also introduce an annotated corpus of aspect categories and sentiment polarities extracted from restaurant reviews in Vietnamese, and conduct a series of experiments on the corpus. Experimental results demonstrate the effectiveness of the proposed approach.

## 1. Introduction

Sentiment analysis and opinion mining [18], the subfield of analysing opinionated text in online product reviews, social networks, blogs, forums, and so on, has become an important and hot research topic in natural language processing (NLP) and data mining (DM). Opinion mining systems provide useful information for not only customers but also service providers and manufactures. For customers, knowing opinions of other users is important in choosing suitable products or services. For service providers and manufacturers, analysing opinions helps in understanding customers, advertising products/services, and choosing strategies for the development of new products.

The most focused and investigated task in the research of sentiment analysis and opinion mining is sentiment classification, which labels a given opinionated text (e.g., a sentence or a customer review) as "positive", "negative", or "neutral". Because sentiment classification assigns an overall sentiment to the input, it is not applicable/suitable for complicated situations. Let us take the following review about a restaurant as an example: "*The staffs are very friendly, but the food is not delicious.*". It is quite difficult to identify the sentiment of the speaker because (s)he satisfies with the restaurant service but complains about the food quality. Some sentiment classification systems assign the "neutral" label to such situations. That information, however, is not so meaningful because we do not know exactly for what (s)he satisfies and/or does not satisfy.

Aspect-based opinion mining (ABOM) [14, 26, 27, 34, 40] addresses the limitation of sentiment classification by recognizing the sentiment on each aspect of products/services expressed in opinionated text. ABOM, therefore, consists of two subtasks: 1) Aspect category extraction, which identifies aspect categories (the entity and attribute pairs) towards which an opinion is expressed in a given text; and 2) Sentiment polarity classification, which assigns a sentiment polarity label to each identified aspect category. For the example in the previous paragraph, the desired output of aspect-based opinion mining process consists of two aspects (service and food quality of the restaurant) and the sentiment polarity on them (positive for service and negative for food quality).

Existing studies [1, 13, 19, 26, 27, 35, 36] on aspect-based opinion mining usually employ supervised learning, the most dominant approach in NLP and DM. Although supervised learning has been shown to be effective in the previous studies [1, 15, 26, 28, 35, 36, 42], it requires to manually annotate a large amount of training data, which is time-consuming and costly. Furthermore, supervised learning methods usually depend on domains. Annotating training data for all domains in every language is impossible. Applying supervised methods to a new domain or a new language, especially a resource-poor language, therefore, is still challenging.

This paper addresses both subtasks of aspect-based opinion mining for Vietnamese, a resource-poor language. To ease the dependence on the amount of annotated data, we introduce a method that utilizes available labelled datasets in foreign languages, especially resource-rich ones, to improve the performance of supervised models. Annotated data in foreign languages are first translated into Vietnamese using a machine translation system. Translated data along with annotated data in Vietnamese language are then consolidated into one dataset, which is used to train aspect category extraction and sentiment polarity classification models. To soften the negative effect of the difference in expressions and words in different languages caused by the data gathering process, we employ a word embedding technique [21], which maps words into a low-dimensional space where semantically similar words are represented by nearby vectors. Word embedding features, therefore, can help classification models recognize synonyms and hypernyms.

Our contributions can be summarized in three main points. First, we propose a method for aspect-based opinion mining in resource-poor languages by exploiting available data in resource-rich languages. Unlike previous work that considers only one subtask [29], here we address both aspect category extraction and sentiment polarity classification by proposing several novel solutions such as automated opinion word extraction and feature generation from neighborhood. The method is general and flexible in the sense that it is language independent and can be used with any supervised learning algorithm. Second, we introduce an annotated corpus for Vietnamese aspect-based opinion mining consisting of 575 reviews with 3796 sentences. We believe that the corpus is a useful resource for the Vietnamese NLP research community. Third, we show the effectiveness of the proposed method by conducting a series of experiments on this corpus. Using translated data from English, our method improves the performance of a Vietnamese aspect-based opinion mining system in both aspect category extraction and sentiment polarity classification subtasks.

The rest of this paper is structured as follows. Section 2 presents related work. Section 3 introduces the proposed method for aspect-based opinion mining in resource-poor languages. In section 4, we present our Vietnamese aspect-based opinion mining corpus and experiments on that corpus. Finally, section 5 concludes the paper and discusses possible directions for future work.

## 2. Related work

In this section we provide a brief review on related work, including methods for aspect category extraction and sentiment polarity classification, and studies on sentiment analysis and opinion mining in Vietnamese.

### 2.1. Methods for Aspect Category Extraction and Sentiment Polarity Classification

Aspect category extraction is the first subtask in aspect-based opinion mining. If we identify a wrong aspect category, the sentiment polarity on that aspect will be meaningless. On the other hand, if we miss an aspect category, we have nothing to do in the second subtask. Extracting the correct list of aspect categories is, therefore, very important. Existing methods for aspect category extraction can be roughly divided into two main approaches *classification-based* and *clustering-based*. Consider two typical situations in aspect-based opinion mining. The first situation, we know exactly the list of aspect categories which we want to extract and we have annotated data with aspect category labels. The second one, we could not define such the list of aspect categories and do not have annotated data.

Methods belonging to the classification-based approach can deal with the first situation by using supervised learning algorithms. Such methods usually model the task as a multi-label classification problem where each label corresponds to an aspect category, or multiple binary classification problems where each problem deal with a specific type of aspect [15, 26, 37].

The clustering-based approach [11] can be used in the second situation. Because we do not have annotated data, unsupervised or semi-supervised learning algorithms with a bootstrapping technique [11, 22, 27] is an appropriate choice. Methods belonging to this approach usually extracts aspect categories in two steps: 1) extracting all aspect terms from a given plain text corpus; and 2) clustering aspect terms with similar meaning into aspect categories. Rule-based and topic modeling, among the others, are two popular methods for aspect term extraction. While a rule-based method employs handcrafted or (semi) automated rules based on syntactic dependency or the relation between opinion and aspect words to extract aspect terms [19], topic modeling methods treat aspect categories like topics, which are distributed over aspect terms in the corpus [4, 5, 22].

Compared to the clustering-based approach, the classification-based one is more accurate because it can utilize supervised learning with annotated data. Our extraction method belongs to the classification-based approach, in which we employ supervised learning algorithms to build multiple binary classifiers. For each aspect category, a binary classifier determines whether the input sentence expresses the aspect category or not. The training data is enriched by leveraging available data from other (resource-rich) languages.

Various methods have been proposed to deal with aspect-based sentiment classification, ranging from traditional learning algorithms to more advanced techniques with neural networks. Wang et al. [36] and He et al. [12] with long short-term memory networks, Xue and Li [38] with convolutional neural networks, Wang et al. [35] and Nguyen and Shirai [23] with recursive networks, and Wang et al. [33] and Majumder et al. [20] with memory networks are examples of successful neural network models for aspect-based sentiment classification, among the others. Deep learning models, however, are usually trained with a large amount of labeled data, which is challenging for natural language processing (NLP) tasks in resource-poor languages.

2.1. Sentiment Analysis and Opinion Mining in Vietnamese

Like in other languages, most of the previous work on sentiment analysis and opinion mining in Vietnamese focuses on sentiment classification. Existing approaches range from rule-based methods [16] to supervised [8, 41] and weakly supervised/semi-supervised learning algorithms [2, 10]. An empirical study on learning-based sentiment classification in Vietnamese is described in Duyen et al. [8]. The authors introduce an annotated corpus in the hotel domain extracted from online reviews and conduct a series of experiments on the corpus. Several feature extraction methods and supervised learning algorithms have been investigated, including Naive Bayes, Maximum Entropy Model, and Support Vector Machines. Ha et al. [10] describe a method using bag-of-bigram features in a lifelong learning framework for cross-domain Vietnamese sentiment classification. Bach and Phuong [2] present a weakly supervised method for sentiment classification in resource-poor languages. The method exploits the overall ratings of reviews as extra information to train a semi-supervised sentiment classifier. Experimental results on two datasets of Japanese and Vietnamese show the effectiveness of their method. Phu et al. [25] propose a valence-totaling model for Vietnamese sentiment classification, which achieves 63.9% accuracy on a corpus consisting of 30,000 Vietnamese documents.

Among previous work on Vietnamese aspect-based opinion mining, Vu et al. [32] present a rule-based method for mining product reviews. Aspect words and opinion words are first extracted using a set of Vietnamese syntactic rules. Customers' opinion orientations and summarization on aspects are then determined by using a dictionary of Vietnamese sentiment words. Le et al. [17] present a semi-supervised method for extracting and classifying aspect terms in Vietnamese text. Their method can be classified into the clustering-based approach. The method first extracts all tokenized words from a plain text corpus and selects the most frequent ones, which are manually labeled and severed as the seed of the algorithm. A decision tree is then built to classify aspect terms. Our work is different from those previous works in that we identify aspect categories directly without extracting aspect terms. Furthermore, our method utilizes existing data from a resource-rich language to improve the task in Vietnamese, a resource-poor language.

Recently, a shared task on Vietnamese aspect-based sentiment analysis has been introduced at the fifth international workshop on Vietnamese Language and Speech Processing (VLSP 2018) [14]. In the context of the shared task, some studies have been presented using supervised learning algorithms [14, 28]. Our work is different from those studies in that we present a new method to solve the task. Furthermore, we identify aspect categories and sentiment polarities in a single sentence instead of a whole review. We believe that the task at the sentence level is more practical and applicable in real world applications.

The idea of using labeled dataset from foreign languages to augment the training dataset in Vietnamese for the task of aspect extraction was first proposed in [29]. In this work, we extend upon this idea by proposing a novel method for classifying sentiment polarity by leveraging labeled data from another language. As will be shown in the next section, the proposed method is not a straightforward modification of [29] but rather come with nontrivial and novel desgin solutions.

## 3. Proposed Method

In this section, we present a method for Vietnamese aspect-based opinion mining from opinionated text, which consists of two subtasks: 1) aspect category extraction; and 2) sentiment polarity classification. For each sentence expressing opinions, the aspect category extraction subtask identifies opinion targets such as product or service features and their categories. In this work, we identify both explicit and implicit aspects. The sentiment polarity classification subtask identifies an opinion or a sentiment polarity (positive, negative, or neutral) for each identified aspect category.

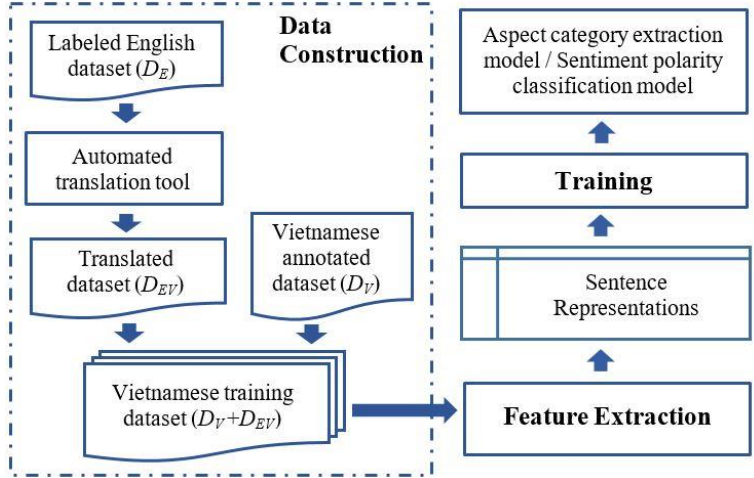

Fig. 1. The proposed framework for Vietnamese aspect-based opinion mining.

For both subtasks, we assume that we have three datasets from the same domain: 1) a **V**ietnamese annotated dataset, denoted by $D_V$, in which each sentence expressing opinions is labeled with aspect categories and sentiment polarities, or "*Null*" if it does not contain any aspect; 2) a labeled dataset in another language (**E**nglish in this work), denoted by $D_E$; and 3) a Vietnamese test dataset, denoted by $T_V$. The $D_E$ dataset is translated into the original language (Vietnamese in this case), denoted by $D_{EV}$, by an automated translation tool. Our goal is to utilize $D_{EV}$ in addition to $D_V$ to improve the accuracy of aspect category extraction and sentiment polarity classification from $T_V$.

The overall proposed framework for solving both subtasks consists of three main steps: 1) data construction for Vietnamese training dataset, 2) feature extraction, and 3) training the aspect category extraction/sentiment polarity classification models. The difference in our proposed method for Vietnamese aspect-based opinion mining is in the first step, data construction. Instead of using only Vietnamese annotated dataset $D_V$ to train the aspect category extraction/sentiment polarity classification models, we supplemented the Vietnamese training dataset with labeled dataset $D_{EV}$, which is translated from English labeled dataset $D_E$ by an automated translation tool. The Figure 1 shows the three steps of the proposed framework.

3.1. Training Dataset Construction

We construct training dataset from two sources: 1) the labeled data in Vietnamese; and 2) the labeled data in foreign language (English in this case). We use the Vietnamese dataset as is. For English dataset, we use Google Translate tool to translate sentences with their aspect categories as well as the corresponding sentiment polarity labels into Vietnamese. The translated labeled dataset $D_{EV}$ is then added to the annotated Vietnamese dataset $D_V$ to form new training dataset ($D_V + D_{EV}$). Actually, we can have a better-quality translation version of the training dataset by translating manually, but this work is very time-consuming and costly. Moreover, manual translation is also difficult to apply to multiple languages when we want to utilize labeled datasets from them.

3.2. Feature Extraction

Before feature extraction process, all Vietnamese sentences, in both original and translated datasets, are segmented into Vietnamese words [31]. This word segmentation is an important preprocessing step in most Vietnamese natural language processing problems because each Vietnamese word may have one or more syllables. The syllables in a Vietnamese word are delimited by white spaces. In the following, we present the feature extraction methods for both aspect category extraction and sentiment polarity classification subtasks.

*a) Aspect Category Extraction*

For this subtask, we extract two kinds of features, namely *basic features* and *word embedding features*, which we use to represent sentences.

*1) Basic features:* Basic features are *n*-grams of Vietnamese words, which are unigrams, bigrams, and trigrams, extracted from segmented Vietnamese sentences. For example, for a given sentence "Đồ_ăn rất ngon." (means "*The food is very delicious.*"), *n*-grams (unigrams, bigrams, and trigrams) of features are extracted as follows: *đồ_ăn, rất, ngon, đồ_ăn rất, rất ngon, đồ_ăn rất ngon*. Although basic features are relatively simple, they have been proven to be effective for most Vietnamese natural language processing tasks.

*2) Word embedding features:* When comparing two sets of words between Vietnamese dataset ($D_V$) and translated dataset ($D_{EV}$) (from English), the result is less than 50% vocabulary overlap. We found that people in different countries often use different styles as well as different words when expressing their ideas. For example, English people often talk "*The food was not great, the waiters were rude.*" (means "Thức_ăn không tuyệt_vời và bồi_bàn thô_lỗ."), but a Vietnamese person may use another expression "Thức_ăn không ngon và bồi_bàn rất mất lịch_sự." (means "*The food was not good and the waiters were very impolite.*"). Therefore, we need to reduce the vocabulary difference between the original and translated texts by extracting several features which can capture the similarity and the relationship of words. For this reason, we propose adding word embedding features to basic features in order to further improve the effectiveness of aspect category extraction process.

Word embeddings [21, 39], a key breakthrough in the research on NLP, are a class of distributed word representation techniques where words are mapped into a low-dimensional vector space. Compared with the traditional word representation method, i.e. one-hot representation, word embeddings have two main advantages. While one-hot vectors are high-dimensional and spare, word embeddings are low-dimensional and dense. Word embeddings, therefore, are more efficient in representation and computation. The second advantage, and more important, is that word embeddings have the ability to generalize due to semantically similar words are represented by close points (similar vectors) in the vector space. Popular method for deriving word embeddings from a large corpus of plain text are dimensionality reduction on the word co-occurrence matrix and neural networks [21, 24, 39], among the others. Word embeddings can be used directly as features in a text/sentence classification task or serve as the first layer in a deep learning architecture [3].

Let $v$ be a function that maps a word $w$ to its word vector representation $v(w)$, word embedding features of a sentence $s$ represented as a sequence of words ($w_1, w_2, \ldots, w_{|s|}$) can be computed as the element-wise sum of the word vectors:

$$(1) \quad v_{em}(s) = \sum_{i=1}^{|s|} v(w_i)$$

where $|s|$ denotes the length of sentence $s$.

*3) Concatenation:* We use a simple method to combine basic features and word embedding features of a sentence $s$ by concatenating two kinds of features as follows:
- Represent $s$ by a one-hot vector $v_{oh}(s)$ of $n$-grams
- Represent $s$ by a word embedding vector $v_{em}(s)$
- Concatenate two vectors $v_{oh}(s)$ and $v_{em}(s)$

*b) Sentiment Polarity Classification*

For the sentiment polarity classification subtask, we employ three main types of features, including *important words*, *word embeddings*, and *aspect category features*, which are described in the following section.

*1) Important words:* Each sentence may have several aspect categories. Therefore, in order to obtain more meaningful features to train a sentiment polarity classification model corresponding to each identified aspect category, we extract words that are important to the aspect category in the sentence. Important words are defined as the ones that identify the aspect category and can be determined through the SVM coefficients when we train the extraction model for that aspect category. The absolute values of the SVM coefficients show how important features (i.e., words) are [9]. We first select top $k$ words (seed words) with the highest absolute values of the SVM coefficients, where *k* is a parameter. In our experiments, *k* was set to 5. To extract sentiment words, we then expand the set of words by selecting related ones using dependency trees. Words that have a relation with a seed word, i.e. head or child, will be selected.

Figure 2 shows the dependency tree of the sentence "*Không_gian siêu đẹp và lãng_mạn.*" (*Space is super beautiful and romantic.*). A relation is represented by a directed link from the head word to the child word and the type of relation is displayed near by the arrow. For example, "*không_gian*" relates to "*đẹp*" by *sub* (subject) relationship and "*siêu*" relates to "*đẹp*" by *amod* (adjectival modifier) relationship. Suppose that "*không_gian*" is a seed word extracted in the first step, then word "*đẹp*" will be selected in the expansion step.

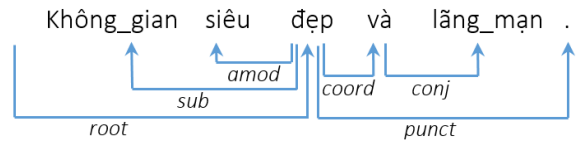

Fig. 2. An example of a dependency tree

*2) Word embedding features:* word embedding features are extracted in a similar way as in aspect extraction. However, we only use important words instead of all words.

*3) Aspect category features:* People can use the similar expressions when commenting about different aspect categories with different opinions. Therefore, a phrase can express a positive opinion about an aspect category but a negative opinion about another one. For example, consider two following sentences: 1) "*Kimchi is not spicy but slightly sweet.*", 2) "*The drink is slightly sweet.*" The phrase "*slightly sweet*" indicates a negative opinion about the food aspect but a positive opinion about the drink aspect. For this reason, aspect category is exploited as a kind of feature for the sentiment polarity classification subtask.

3.2. Training Models

Let *N* denote the number of aspect categories that we want to extract. For the aspect category extraction subtask, we train a binary classifier for each aspect category to predict whether or not a sentence contains the aspect category. For the sentiment polarity classification subtask, we train a classifier to predict which sentiment polarity (positive, negative, neutral) of an identified aspect category in the sentence. Totally, we have *N* classifiers for *N* aspect categories, and one classifier for identifying sentiment polarities.

While any supervised learning algorithm can be used in our framework, we chose Support Vector Machines (SVMs) [30] due to their effectiveness for various classification tasks in NLP [8, 15, 26, 28]. SVMs are based on two main principles.

First, SVMs separate samples with different labels by a hyperplane so that the distance from the hyperplane to the samples with different labels is the largest. This principle is called maximum margin. In the training process, an SVM algorithm determines a hyperplane with maximum margins by solving the optimal problem for a quadratic objective function. Second, to solve the case of non-separation sample by a hyperplane, the SVM method maps the original space of the samples to a higher dimensional space, then find the hyperplane with maximum margin in this new space. In order to increase the performance of the mapping, a technique used is kernel function, for example, polynomial kernel function, Gaussian radial basic kernel function.

## 4. Experiment Results and Analysis

To verify the effectiveness of the proposed method, we built an annotated Vietnamese dataset and conducted a series of experiments on it. We exploited English dataset from SemEval-2016 Task 5 [26] as the supplemented data. In the following we describe two datasets, i.e. Vietnamese dataset and English dataset, and experimental results as well as analysis.

4.1. Datasets

*a) Vietnamese Dataset*

To construct our Vietnamese dataset, we perform three steps as follows: 1) raw data collection, 2) preprocessing, and 3) data annotation.

*1) Raw data collection:* We collected raw data from Foody website (at *https://www.foody.vn/*). This is one of the most popular and the largest websites in Vietnam, where people can search, evaluate, comment, and book not only for food and drink but also for other services such as travel, beauty, entertainment, shopping, and so on. Raw data was extracted from reviews of several restaurants located in Vietnam (most of them in Hanoi and Ho Chi Minh city).

*2) Preprocessing:* We conducted several preprocessing steps on extracted raw data, consisting of data cleaning and sentence boundary detection. As a result, we obtained 575 reviews including 3796 Vietnamese sentences.

*3) Data annotation:* We asked three annotators to assign aspect categories and their sentiment polarities for preprocessed sentences. These annotators are computer science students with basic background on machine learning and natural language processing, in which two annotators are undergraduates and the third one is a postgraduate. First, the two undergraduates assigned labels independently and then the postgraduate examined and made the final decision if there was any disagreed assignment from two previous annotators. We measured the inter-annotator agreement by Cohen's kappa coefficient.

In this work, we computed two Cohen's kappa scores for two types of labels, i.e. aspect categories and sentiment polarities. Because each sentence can be labeled by multiple aspect categories and sentiment polarities, we computed the Cohen's kappa coefficient for each type of category and sentiment polarity, and reported the average scores. The Cohen's kappa scores of our Vietnamese dataset are 0.83 for aspect categories and 0.86 for sentiment polarities, which can be interpreted as almost perfect agreement [7].

As in SemEval-2016 Task 5 [26], we assigned 12 aspect category labels represented by 12 tuples of (entity, attribute).

Table 1. Statistics of two datasets

|  | Vietnamese | English | Total |
|---|---|---|---|
| Number of reviews | 575 | 440 | 1015 |
| Number of sentences | 3796 | 2676 | 6472 |

Table 2. Aspect categories and their sentiment polarities in two datasets

| Aspect Category | Vietnamese | | | | English | | | |
|---|---|---|---|---|---|---|---|---|
|  | *Pos.* | *Neg.* | *Neu.* | *Total* | *Pos.* | *Neg.* | *Neu.* | *Total* |
| RESTAURANT#GENERAL | 178 | 30 | 25 | **233** | 420 | 135 | 9 | **564** |
| RESTAURANT#PRICES | 71 | 44 | 17 | **132** | 40 | 53 | 8 | **101** |
| RESTAURANT#MISCELLANEOUS | 157 | 22 | 15 | **194** | 74 | 40 | 17 | **131** |
| FOOD#QUALITY | 957 | 247 | 153 | **1357** | 886 | 235 | 41 | **1162** |
| FOOD#STYLE_OPTIONS | 475 | 80 | 31 | **586** | 114 | 60 | 18 | **192** |
| FOOD#PRICES | 115 | 76 | 16 | **207** | 47 | 62 | 4 | **113** |
| DRINKS#QUALITY | 213 | 56 | 38 | **307** | 61 | 6 | 2 | **69** |
| DRINKS#STYLE_OPTIONS | 56 | 14 | 5 | **75** | 40 | 4 | 0 | **44** |
| DRINKS#PRICES | 29 | 19 | 8 | **56** | 13 | 11 | 0 | **24** |
| SERVICE#GENERAL | 345 | 100 | 42 | **487** | 283 | 302 | 19 | **604** |
| AMBIENCE#GENERAL | 371 | 92 | 53 | **516** | 258 | 44 | 19 | **321** |
| LOCATION#GENERAL | 95 | 101 | 19 | **215** | 32 | 1 | 8 | **41** |
| **Total** | **3062** | **881** | **422** | **4365** | **2268** | **953** | **145** | **3366** |

*b) English Dataset*

English dataset was retrieved from SemEval-2016 Task 5 [26] with both training set and gold test set to use as the additional resource in our proposed method. We obtained English dataset including 440 reviews with 2676 sentences. Finally, in total, we had 1015 reviews with 6472 sentences for both Vietnamese and English datasets. Table 1 presents the statistical information of two datasets.

In Table 2, we show 12 aspect categories and their sentiment polarities (positive, negative and neutral) in detail in our Vietnamese dataset and the English dataset. In both datasets, the highest frequency aspect category is FOOD#QUALITY (with

1357 and 1162 times corresponding to Vietnamese and English datasets), while the lowest frequency aspect categories include DRINKS#STYLE OPTIONS and DRINKS#PRICES (with under 80 times for each aspect category). And there is one thing to note that the number of positive samples is relatively higher than the number of negative samples in both datasets.

4.2. Experimental Setup

For both subtasks, we randomly divided the Vietnamese dataset into 10 folds and conducted cross-validation tests. The performances of aspect category extraction and sentiment polarity classification models were measured using the *precision*, *recall*, and the $F_1$ score on each type of aspect category and each type of sentiment polarity label (positive and negative).

$$Precision = \frac{|A \cap B|}{|A|} \quad (2)$$

$$Recall = \frac{|A \cap B|}{|B|} \quad (3)$$

$$F_1 = \frac{2 * Precision * Recall}{Precision + Recall} \quad (4)$$

- **Aspect category extraction:** Let's consider aspect category RESTAURANT#GENERAL (general comment on the restaurant) as an example. The *precision*, *recall*, and the $F_1$ score for this aspect category can be computed as in (2), (3) and (4). In which, *A* denotes the set of sentences with aspect category RESTAURANT#GENERAL identified by the model, and *B* denotes the set of sentences with this aspect category labeled by human. It is similar for other aspect categories.

- **Sentiment polarity classification:** Let's consider aspect category RESTAURANT#GENERAL. We consider sentiment polarity *Positive* for this aspect category as an example. The *precision*, *recall*, and the $F_1$ score for this sentiment polarity of aspect category RESTAURANT#GENERAL can be computed as in (2), (3) and (4). In which, *A* denotes the set of sentences with positive aspect category RESTAURANT#GENERAL identified by the model, and *B* denotes the set of sentences with positive aspect category RESTAURANT#GENERAL labeled by human. It is similar for negative sentiment polarity of this aspect category, and similar for other aspect categories which we want to extract.

In our experiments, all sentences in the English dataset were translated into Vietnamese using Google Translate (*https://translate.google.com/*). For word embeddings, we used 50-dimensional word vectors trained with Word2Vec [21] on a text collection from Baomoi (*https://baomoi.com/*) with more than 433,000 Vietnamese words.

4.3. Baselines

For both subtasks, aspect category extraction and sentiment polarity classification, we conducted experiments to compare the performance of three models (summarized in Table 3) consisting of baseline model, CRL model (Cross-Language), and WEmb model (Word Embedding). All models were trained with linear SVMs [6].

Table 3. Models to compare

| Model | Data | Features | |
|---|---|---|---|
| | | *Subtask1* | *Subtask2* |
| Baseline | Vietnamese | *n*-grams | Important words + Aspect category |
| CRL | Vietnamese + English | *n*-grams | Important words + Aspect category |
| WEmb | Vietnamese + English | *n*-grams + Word Embeddings | Important words + Word Embeddings + Aspect category |

- **Baseline Model:** This model used only the Vietnamese dataset. We used *n*-grams features for aspect category extraction, and important words and aspect category features for sentiment polarity classification. The purpose was to investigate the effectiveness of our proposed method when using only the Vietnamese dataset with basic features.

- **CRL Model (Cr**oss-**L**anguage Model**):** We used both sentences in the Vietnamese dataset and the translated sentences from the English dataset to train the model. The feature sets for two subtasks were similar to the ones of the baseline model (i.e., *n*-grams features for aspect category extraction, and important words and aspect category features for sentiment polarity classification). The purpose of this experiment was to investigate the impact of using additional translated data from another language on the aspect category extraction and sentiment polarity classification subtasks.

- **WEmb Model (W**ord **Emb**edding Model**):** This model was similar to our CRL model but adding word embedding features. It means that we used both Vietnamese dataset and translated dataset from English as training dataset; and the feature sets consisting of *n*-grams and word embedding features for aspect category extraction, and important words, word embedding and aspect category features for sentiment polarity classification. The purpose was to investigate the effectiveness of word embedding features on the performance of the aspect category extraction and sentiment polarity classification models.

4.4. Experimental Results

*a) Results on Aspect Category Extraction*

First, we conducted experiments with the baseline model. We investigated three variants of the feature set: 1) unigrams only; 2) unigrams and bigrams; and 3) unigrams, bigrams, and trigrams. The obtained results showed that the variant with unigrams and bigrams is the best. For this reason, in this work, we report the experimental results of the second variant and skip over other variants.

In Table 4, we present the experimental results using unigrams and bigrams on aspect category extraction of baseline model. The most accurate aspect categories achieved about 80% in the $F_1$ score, consisting of three categories, SERVICE#GENERAL, FOOD#QUALITY, and AMBIENCE#GENERAL with 83.52%, 80.40%, and 79.58%, respectively. In the corpus, all three categories appear with higher frequency compared to most of the other aspect categories: 487 times for SERVICE#GENERAL, 1357 times for FOOD#QUALITY, and 516 times for AMBIENCE#GENERAL. Aspect categories with the lowest $F_1$ scores were DRINKS#STYLE OPTIONS and DRINKS#PRICES with 5.86% and 16.27%, respectively. These two aspect categories appear with very small frequency: 44 times for DRINKS#STYLE OPTIONS and 24 times for DRINKS#PRICES. The aspect category LOCATION#GENERAL is an interesting case. Although this aspect category appears with a relatively few frequency (215 times), the result in the $F_1$ score was impressive with 76.90%. There are two reasons that may explain for this: 1) there is only one aspect category for location in the corpus, and 2) sentences describe location usually containing several specific phrases such as "*nơi*"/"*vị_trí*" (location), "*đặt tại*"/"*đặt vị_trí tại*" (located at), "*trung_tâm*" (center), "*phố*" (street), and "*đường*" (road). On average, the baseline model achieved 78.00%, 64.52%, and 70.62% in *precision*, *recall*, and the $F_1$ score, respectively.

We then performed experiments in order to compare our proposed models with the baseline model. We present experimental results (in the $F_1$ score) of our CRL and WEmb models in the Table 4. The results show that CRL model won in 9 out of 12 aspect categories compared with the baseline model that proving the effectiveness of supplementing translated data for the aspect extraction subtask. Several aspect categories with a clear improvement in the $F_1$ score consist of RESTAURANT#GENERAL (10.46%), RESTAURANT#PRICES (8.58%), DRINKS#STYLE_OPTIONS (5.00%), and FOOD#PRICES (3.13%). On average, our CRL model achieved an $F_1$ score of 71.77%, which improved 1.15% compared with the baseline model. Moreover, by adding word embedding features to train WEmb model, we obtained better results in 9 out of 12 aspect categories compared with CRL model. Some aspect categories got the clear improvement results in the $F_1$ score are DRINKS#PRICES (7.69%), LOCATION#GENERAL (2.67%), DRINKS#QUALITY (1.8%), and AMBIENCE#GENERAL (1.13%). On average,

WEmb model retrieved 72.33% in the $F_1$ score, which improved 1.71% and 0.56% compared with the baseline model and CRL model, respectively.

Table 4. Experimental results on aspect category extraction of our proposed models

| Aspect Category | Baseline | | | CRL | WEmb |
|---|---|---|---|---|---|
| | *Pre. (%)* | *Rec. (%)* | *$F_1$ (%)* | *$F_1$ (%)* | *$F_1$ (%)* |
| RESTAURANT#GENERAL | 53.97 | 35.53 | 42.20 | 52.66 | 53.48 |
| RESTAURANT#PRICES | 52.57 | 33.53 | 40.39 | 48.97 | 49.83 |
| RESTAURANT#MISCELLANEOUS | 76.85 | 57.65 | 64.17 | 61.84 | 61.19 |
| FOOD#QUALITY | 82.95 | 78.42 | 80.40 | 80.47 | 80.99 |
| FOOD#STYLE_OPTIONS | 79.82 | 64.51 | 69.32 | 68.90 | 68.45 |
| FOOD#PRICES | 50.55 | 30.61 | 36.85 | 39.98 | 37.21 |
| DRINKS#QUALITY | 72.65 | 56.03 | 62.20 | 61.38 | 63.18 |
| DRINKS#STYLE_OPTIONS | 15.00 | 4.11 | 5.86 | 10.86 | 11.81 |
| DRINKS#PRICES | 33.33 | 11.00 | 16.27 | 16.52 | 24.21 |
| SERVICE#GENERAL | 93.32 | 76.00 | 83.52 | 84.30 | 84.78 |
| AMBIENCE#GENERAL | 89.43 | 72.54 | 79.58 | 80.46 | 81.59 |
| LOCATION#GENERAL | 90.07 | 68.26 | 76.90 | 77.09 | 79.76 |
| **Average** | **78.00** | **64.52** | **70.62** | **71.77** | **72.33** |

*b) Results on Sentiment Polarity Classification*

Table 5 shows experimental results of three models on the sentiment polarity classification subtask. Our first observation is that for all models the $F_1$ score of the positive class was much higher than that of the negative class: 81.45% vs. 47.33% (Baseline model), 83.43% vs. 48.20% (CRL model), and 83.63% vs. 50.19% (WEmb model). One reason is the dominance of positive samples in the datasets, 5330 positive samples compared with 1834 negative ones. Another reason may be that positive opinions are usually stated directly and explicitly while negative opinions are often implicit. For example, it is quite easy to determine the positive sentiment in sentence "*We are very satisfied with the food of this restaurant.*" because the sentence contains a strong positive indication word "*satisfied*". However, it is more difficult to identify negative sentiments in sentences "*We had to wait for food about half an hour.*" and "*Kimchi is not spicy but slightly sweet.*" because they seem to be stated implicitly.

The second observation is that two proposed models, CRL and WEmb, outperformed the baseline model on both positive and negative classes. For CRL model, we achieved the $F_1$ scores of 83.43% (for positive class) and 48.20% (for negative class), which improved 1.98% (positive) and 0.87% (negative). By adding word embedding features, WEmb model got the highest $F_1$ scores on both classes with 83.63% (positive) and 50.19% (negative), which improved 2.18% and 2.86% compared with the baseline model. The results showed the effectiveness of utilizing the translated English dataset and word embedding features for Vietnamese sentiment polarity classification.

Table 5. Experimental results on sentiment classification (with k = 5 words)

| Sentiment Polarity | Baseline | | | CLR | | | WEmb | | |
|---|---|---|---|---|---|---|---|---|---|
| | *Pre.* | *Rec.* | *$F_1$* | *Pre.* | *Rec.* | *$F_1$* | *Pre.* | *Rec.* | *$F_1$* |
| Positive | 82.97 | 80.09 | 81.45 | 80.99 | 86.12 | 83.43 | 81.55 | 85.93 | **83.63** |
| Negative | 45.23 | 50.27 | 47.33 | 49.52 | 47.56 | 48.20 | 50.86 | 50.17 | **50.19** |

In the above experiments, we train a single sentiment classifier for all aspect categories in which the aspect information serves as features for the model. The advantage of this method is that we can exploit all the available training samples, regardless of their aspect categories. In the next experiments, we try to build multiple sentiment classifiers, one for each aspect category. Though this method limits the amount of training samples, each classifier focuses on a specific aspect category. Table 6 shows the $F_1$ scores of three models using the new training strategy. For the positive class, all models achieved good results (higher than 80%) on most aspect categories except for DRINKS#PRICES. Compared to the baseline model, CRL and WEmb got better results on some aspect categories, including RESTAURANT#PRICES, RESTAURANT#MISCELLANEOUS, FOOD#STYLE_OPTIONS, FOOD#QUALITY, DRINKS#STYLE_OPTIONS, and SERVICE#GENERAL.

Table 6. The F1 scores of sentiment polarity classification (one classifier for each aspect category), with k = 5 words

| Aspect Category | Baseline | | CRL | | WEmb | |
|---|---|---|---|---|---|---|
| | *Pos.* | *Neg.* | *Pos.* | *Neg.* | *Pos.* | *Neg.* |
| RESTAURANT#GENERAL | 88.12 | 32.00 | 87.51 | 36.38 | 85.69 | 32.97 |
| RESTAURANT#PRICES | 84.11 | 70.86 | 84.45 | 65.65 | 84.57 | 66.44 |
| RESTAURANT#MISCELLANEOUS | 92.81 | 28.33 | 91.32 | 26.33 | 92.14 | 31.38 |
| FOOD#QUALITY | 87.30 | 48.44 | 85.75 | 45.65 | 85.64 | 51.35 |
| FOOD#STYLE_OPTIONS | 90.53 | 36.86 | 90.95 | 45.22 | 89.51 | 37.80 |
| FOOD#PRICES | 85.40 | 75.78 | 86.34 | 76.84 | 79.98 | 69.61 |
| DRINKS#QUALITY | 84.74 | 35.81 | 83.39 | 24.38 | 83.23 | 30.55 |
| DRINKS#STYLE_OPTIONS | 85.61 | 10.00 | 87.79 | 10.00 | 86.93 | 20.00 |
| DRINKS#PRICES | 72.29 | 40.00 | 77.40 | 53.00 | 67.85 | 48.33 |
| SERVICE#GENERAL | 84.93 | 47.38 | 87.05 | 59.69 | 86.99 | 61.09 |
| AMBIENCE#GENERAL | 89.03 | 51.55 | 89.03 | 50.60 | 88.15 | 51.66 |
| LOCATION#GENERAL | 90.24 | 90.59 | 86.36 | 85.00 | 83.71 | 82.17 |

## 5. Conclusion and Future Work

In this paper, we have presented a classification-based method for aspect-based opinion mining in Vietnamese, a resource-poor language, by leveraging available labeled datasets from foreign (resource-rich) languages to improve the effectiveness of supervised classifiers. The method is general and flexible in the sense that it is language independent and can be used with any supervised learning algorithm. Experimental results on our Vietnamese annotated corpus (in the restaurant domain) for aspect-based opinion mining showed that enriching the training dataset with translated English data would greatly increase the performance of the aspect category extraction and sentiment polarity classification models. In addition, using word embedding features further improved the effectiveness of the aspect category extraction and sentiment classification processes.

We plan to study other methods for utilizing available data from resource-rich languages to improve NLP tasks in resource-poor languages, specifically in Vietnamese language. We focus on two directions: 1) leveraging multiple datasets of the same NLP task from different resource-rich languages to deal with the task in Vietnamese language; and 2) leveraging multiple datasets of different NLP tasks from a resource-rich language to solve those tasks in Vietnamese language (i.e. multi-task learning).